\documentclass[]{article}
\usepackage{authblk}
\usepackage{fullpage}

\usepackage{multirow}
\usepackage{caption}
\usepackage{Definitions}
\usepackage[english]{babel}
\usepackage[latin1]{inputenc}
\usepackage{amsfonts}
\usepackage{amsmath}
\usepackage{latexsym}
\usepackage{hyperref}
\usepackage{mathrsfs}
\usepackage{etoolbox}
\usepackage{enumitem}
\usepackage{soul}
\usepackage{stackengine}
\usepackage{graphicx}
  \usepackage{subfig}
\usepackage[vlined,ruled]{algorithm2e}
\usepackage{algorithmic}
\usepackage{colortbl}
\usepackage{algorithmic}
\definecolor{LightBlue}{rgb}{0.90,0.98,1}
\definecolor{LightOrange}{rgb}{1,0.84,0.80}
\definecolor{LightRed}{rgb}{1,0.50,0.50}
\definecolor{LightGrey}{rgb}{0.9,0.9,0.9}
\definecolor{Green}{rgb}{0.55,0.70,0.0}
\definecolor{LightGreen}{rgb}{0.72,0.91,0.80}
\definecolor{LightYellow}{rgb}{0.94,0.98,0.85}
\newcommand{\ord}[1]{\cellcolor{LightGrey}{#1}}
\newcommand{\best}[1]{\cellcolor{LightGreen}{#1}}
\newcommand{\secBest}[1]{\cellcolor{LightBlue}{#1}}

\newcommand{\bad}[1]{\cellcolor{LightOrange}{#1}}
\newcommand{\xhdr}[1]{\vspace{1mm}\noindent{{\bf #1.}}}
\RequirePackage{enumitem}
\setlist{itemsep=3pt,topsep=4pt,parsep=3pt}
\RequirePackage[textsize=scriptsize, disable]{todonotes}
\DeclareMathOperator{\CMI}{CMI}
\newcommand{\ourmodel}{VACS}
\def\SOS{\texttt{SOS}}
\def\EOS{\texttt{EOS}}
\def\rqc{r_{q,c}}
\def\hqc{h^{(q,c)}}
\def\rql{r_{q,l}}
\def\hql{h^{(q,l)}}
\def\rpl{r_{p,l}}
\def\hpl{h^{(p,l)}}
\def\rpc{r_{p,c}}
\def\hpc{h^{(p,c)}}
\def\fqc{f_{q,c}}
\def\fql{f_{q,l}}
\def\fpc{f_{p,c}}
\def\fpl{f_{p,l}}

\newcommand{\niloy}[1]{{\color{red} \bf [NG: #1]}}
\newcommand{\bsblue}[1]{{\color{blue} [#1]}}

\newcommand\blfootnote[1]{%
  \begingroup
  \renewcommand\thefootnote{}\footnote{#1}%
  \addtocounter{footnote}{-1}%
  \endgroup
}
\begin{document}
\title{A Deep Generative Model for Code-Switched Text}

\author[1]{Bidisha Samanta$^{*}$}
\author[1]{Sharmila Reddy$^{*}$}
\author[1]{Hussain Jagirdar}
\author[1]{Niloy Ganguly}
\author[2]{Soumen Chakrabarti}

\affil[1]{Indian Institute of Technology, Kharagpur, \{bidisha, sharmilanangi, hussainjagirdar.hj\}@iitkgp.ac.in, niloy@cse.iitkgp.ernet.in}
\affil[2]{Indian Institute of Technology, Bombay, soumen@cse.iitb.ac.in}

\date{}



\maketitle
\blfootnote{$^{*}$Both the authors contributed equally}
\pagestyle{plain} \thispagestyle{plain}

\begin{abstract}

Code-switching, the interleaving of two or more languages within a sentence or discourse is pervasive in multilingual societies. Accurate language models for code-switched text are critical for NLP tasks. State-of-the-art data-intensive neural language models are difficult to train well from scarce language-labeled code-switched text. A potential solution is to use deep generative models to synthesize large volumes of realistic code-switched text. Although generative adversarial networks and variational autoencoders can synthesize plausible monolingual text from continuous latent space, they cannot adequately address code-switched text, owing to their informal style and complex interplay between the constituent languages. We introduce \ourmodel, a novel variational autoencoder architecture specifically tailored to code-switching phenomena. \ourmodel{} encodes to and decodes from a two-level hierarchical representation, which models syntactic contextual signals in the lower level, and language switching signals in the upper layer. Sampling representations from the prior and decoding them produced well-formed, diverse code-switched sentences. Extensive experiments show that using
synthetic code-switched text with 
natural monolingual data
results 
in significant (33.06\%) drop in perplexity.
\end{abstract}


\section{Introduction}
\label{sec:Intro}


Multilingual text is very common on social media platforms like Twitter and Facebook. A prominent expression of multilingualism in informal text and speech is \emph{code-switching}: alternating between two languages, often with one rendered in the other's character set. Many NLP tasks benefit from accurate statistical language models. Therefore, extending monolingual language models to code-switched text is important.

Many state-of-the-art monolingual models are based on recurrent neural networks (RNNs) \cite{chandu2018language,garg2018code,winata2018code}. We call them RNN language models or RNNlms. RNNlm decoders, conditioned on task-specific features, are heavily used in machine translation \cite{sutskever2014sequence,bahdanau2014neural}, 
image captioning
\cite{vinyals2015show,donahue2015long}
textual entailment \cite{bowman2015large} 
and speech recognition~\cite{chorowski2015attention}.

Training RNNlms is data-intensive. The paucity of language-tagged code-switched text has been a major impediment to training RNNlms well. This strongly motivates the automatic generation of plausible synthetic code-switched text to train state-of-the-art neural language models.

Synthetic but realistic monolingual text generation is itself a challenging problem, on which recent deep generative techniques have made considerable progress.
Two generative architectures are predominantly used: (a)~Generative Adversarial Networks (GAN) \cite{goodfellow2014generative} and (b)~Variational AutoEncoders (VAE) \cite{kingma2013auto}. Several recent works have successfully extended GANs
\cite{zhang2017adversarial,kannan2017adversarial}
and VAEs \cite{bowman2015generating} to generate diverse and plausible synthetic monolingual texts.

Generating plausible code-switched text is an even more delicate task than generating monolingual text. Linguistic studies show that bilingual speakers switch languages by following various complex constraints \cite{myers1997duelling,muysken2000bilingual} which may even include the intensity of sentiment expressed in various segments of text~\cite{rudra2016understanding}. \cite{Pratapa:2018:ACL} synthesized code-mixed sentences by leveraging linguistic constraints arising from Equivalence Constraint Theory. \begingroup 
While this works well for language pairs with good structural correspondence (like English-Spanish), we observe performance 
degrades with weaker correspondence (like English-Hindi). \endgroup ~\cite{bidishaACL} proposes a method to generate code-switched text given a source and target sentence pair, however they can only generate restrictive set of switching patterns. Therefore, a code-switched text synthesizer needs to learn overall syntax distributions of code-switched sentences, as well as model complex switching patterns conditioned on it.

Owing to its great syntactic and switching diversity, large volumes of language-labeled code-switched text is needed to train monolingual deep generative models, which are not available. The only alternative is to train monolingual models with parallel corpora of the two constituent languages which may be relatively easily obtainable. However, training a GAN with aligned parallel corpora may not help, because it is designed to generate a sentence from a noise distribution instead of any learned latent embedding space. Using a VAE RNNlm~\cite{bowman2015generating} is more promising. Aligned parallel corpora are expected to yield similar representations for a source-target sentence pair. Therefore, a VAE decoder conditioned on this embedding may generate some code-switch text without applying explicit external force. However,
it is unlikely to learn subtle connections between context and switching decisions as well as a customized VAE solution, which is our goal.


Here we present \ourmodel, a new deep generator of code-switched text, based on a hierarchical VAE augmented with language- and syntax-informed switching components. 
\begin{itemize}[leftmargin=*]
\item Observed language-labeled code-switched text is encoded to a two-layer compressed representation. The lower layer encodes sequential word context. Conditioned on this lower layer, the encoder models the switching behavior in the higher layer. This contrasts with existing dual-RNNlm architectures \cite{garg2018code} that do not have any explicit gadget to model the switching behavior. Our encoder learns the two-layer representation via variational inference so that the resulting encoded representations enable our decoder to readily generate new code-mixed sentences.

\item Our decoder is designed to sample a context sequence, given a switching pattern. Unlike previous RNNlms \cite{chandu2018language} which consider context and tag generation as independent processes, the decoder of \ourmodel\ first decodes a switching pattern from the switching embedding and then uses this switching pattern memory as well as the lower-layer compressed encoding, to generate a context sequence. The restricted word sequence sampling space improves output quality.

\item During the decoding process, \ourmodel\ (trivially) generates the language labels for each word in the sentence. Thus, VACS lets us synthesize unlimited amounts of labeled code-switched text, starting with modest-sized samples.
\end{itemize}
Owing to the asymmetry between word and label sequences, our encoder and decoder layers show some asymmetries tailored to code-switching, which distinguishes \ourmodel{} from a regular RNN-based VAE.


Through extensive experiments reported here, we establish that augmenting scarce natural labeled code-switched text with plentiful synthetic code-switched text generated by \ourmodel{} significantly improves the 
perplexity of state-of-the-art language models. The perplexity of the models on 
\begingroup 
held-out natural \endgroup Hindi-English text drops by 
33.32\% compared to using only natural training data. Manual inspection also shows that \ourmodel{} generates sentences with diverse mixing patterns.\footnote{https://github.com/bidishasamantakgp/VACS} 

\section{Background on VAEs}
\label{sec:Background}

VAEs \cite{kingma2013auto} are among the most popular deep generative models. They define a \textbf{decoding} probability distribution $p_\theta (x|z)$ to generate observation $x$, given latent variables $z$, which are sampled from a simple \textbf{prior} distribution $p_\pi(z)$. The objective of the VAE is to learn an approximate probabilistic inference model $q_\phi(z|x)$ that \textbf{encodes} latent factors or features $z$ of the variation in the observed data~$x$.

Distributions $p$ and $q$ are often parameterized using deep neural networks. 
We use the maximum likelihood principle to train a VAE, i.e., maximize the expected lower bound of the likelihood on 
observed data $x \sim D$:
\begin{align*}
 \max_\phi 
 \max_\theta 
 \EE_D \Bigl[ \EE_{q_{\phi}(z | x)}\log p _{\theta}(x|z)
 -\text{KL}\bigl(q_\phi(z|x)||p_\pi(z)\bigr) \Bigr]
\end{align*}
To represent more complex features in the latent space, multiple VAEs are stacked hierarchically \cite{rezende2014stochastic,sonderby2016ladder}. The stack of latent variables $Z$ are designed to learn a ``feature hierarchy". For a hierarchical VAE with $\Lambda$ layers, the prior, encoding and decoding probability distributions are modeled as below:
\begin{align}
\text{Encoder:} && q_\phi(Z|x) &= q_\phi(z_1|x) \textstyle
\prod_{\lambda=2}^{\Lambda} q_\phi(z_\lambda|z_{\lambda-1}) \notag \\
\text{Prior:} && p_\pi(Z) &= p(z_\Lambda) \textstyle 
\prod_{\lambda=1}^{\Lambda-1} p_\theta(z_\lambda|z_{\lambda+1}) \label{eq:EPD} \\ 
\text{Decoder:} && p_\theta(x|Z) &= p_\theta(x|z_1) \notag
\end{align}
The performance of the above scheme is sensitive to the design of the layers. Layers $\lambda\gg1$ may fail to capture extra information. Excessively deep hierarchies with large $\Lambda$ may lead to training difficulties~\cite{sonderby2016ladder}.

\section{\ourmodel: A VAE for code-switched text}
\label{sec:OurModel}

This section gives a high-level overview of \ourmodel, followed by details of the building blocks, highlighting key advances beyond prior art. Our focus will be on components that implement a context-based switching distribution. Later, we describe the training process and other implementation details.

\begin{figure*}[ht]
\centering
\subfloat[Encoder (inference)]{\includegraphics[trim={0 6.2cm 0 1.95 cm}, clip,width=.5\columnwidth]{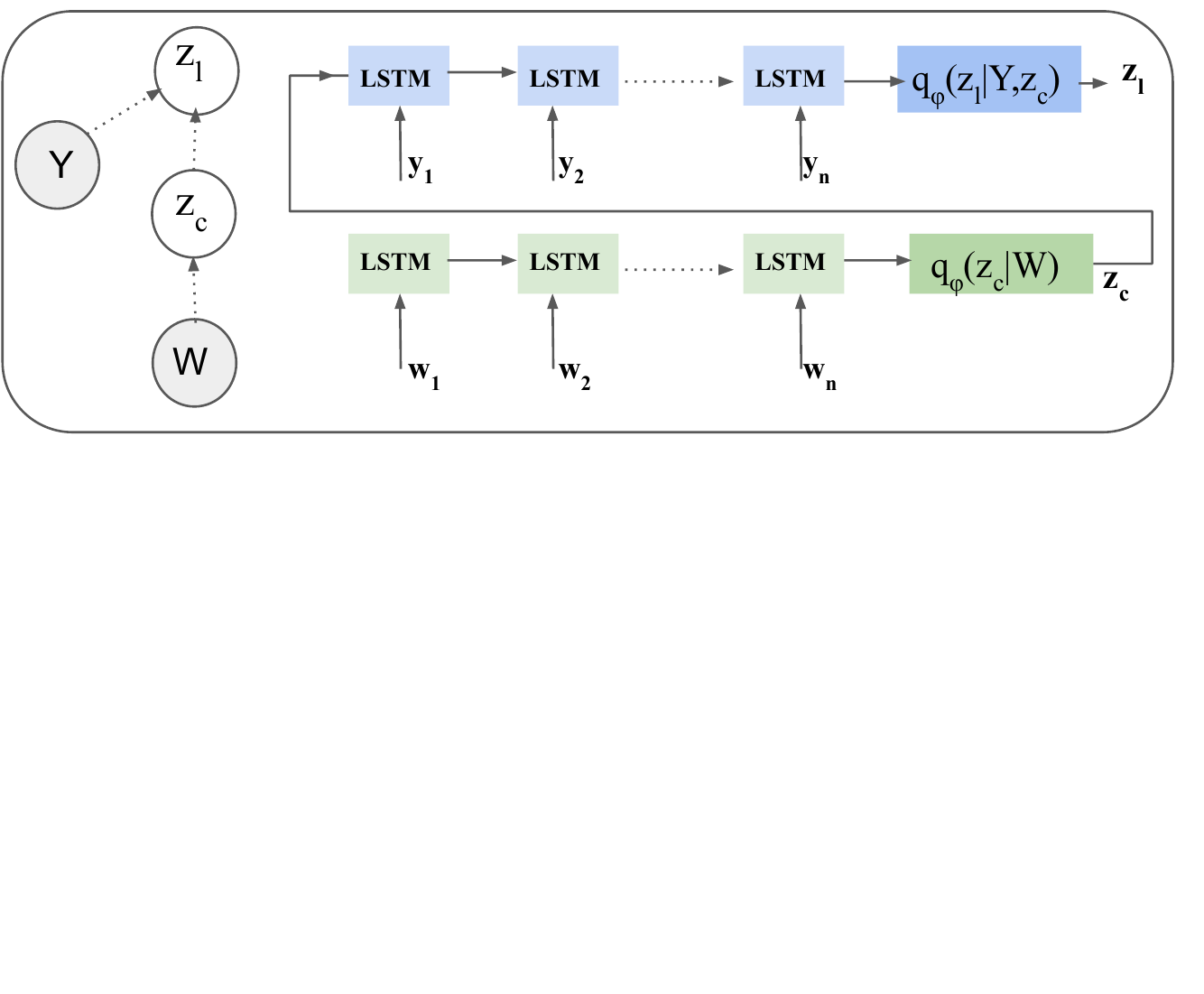}}
\subfloat[Decoder (generation)]{\includegraphics[clip,width=.5\columnwidth]{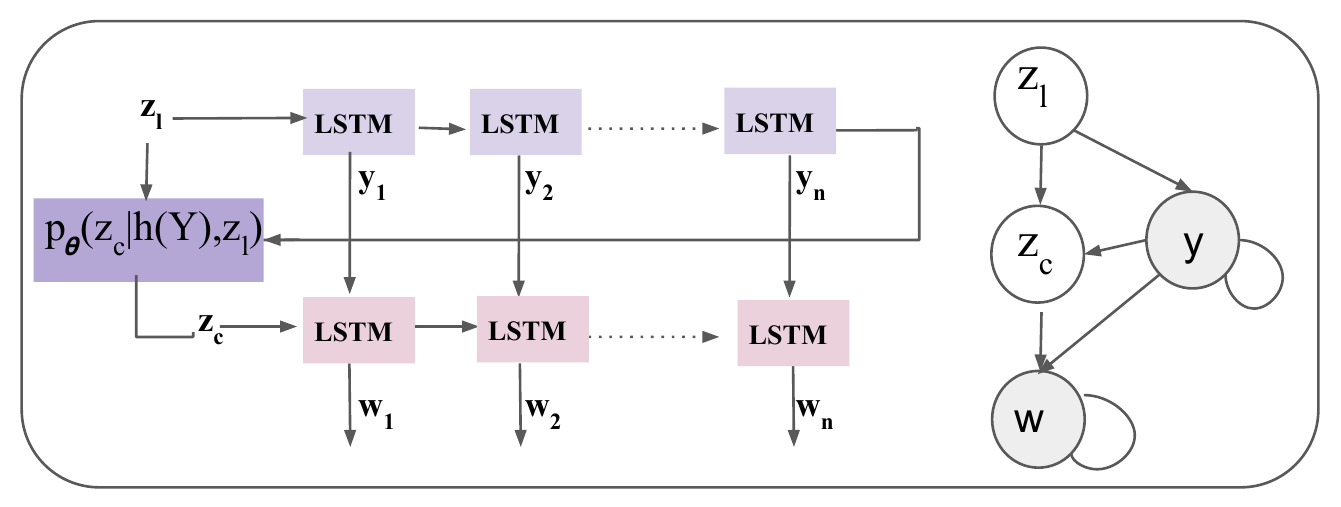}}
\caption{The encoder and decoder in \ourmodel{}. (a)~Graphical model and the recurrent architecture of the encoder. (b)~Graphical model and recurrent architecture of the decoder. 
}
\label{fig:encoderdecoder}
\end{figure*}

\subsection{Overview}
\label{sec:OurModel:Overview}
We aim to design a VAE for code-switched text, which, once trained on a collection of 
tagged code-mixed text should be able to generate new code-mixed text from the same vocabulary.
We represent a code-switched sentence $S$ as $\{(w_i, y_i): i=1,\ldots,|S|\}$, where $(w_i, y_i)$ is a pair comprised of a word $w_i$ at position $i$ and the corresponding language label $y_i$ to which it belongs. Here we consider the simple case of switching between two languages, a source language $s$ and a target language~$t$.
Let \SOS, \EOS{} denote start and end of sentence markers. 
Generation of output stops when label $\EOS$ is generated.
We let $y_i$ take values from $\{s, t, \EOS \}$.
When discrete values like $w_i, y_i$ are input to networks, they are one-hot encoded.
\ourmodel{} is characterized by these components:
\begin{description}
\item[Prior:] $p_\pi(Z)$
\item[Inference model (encoder):] $q_{\phi}(Z | \Wb, \Yb)$
\item[Generative model (decoder):] $p_{\theta}(\Wb, \Yb | Z)$
\end{description}
In our formulation, $Z$ will consist of two latent encoded representations $z_l$ and $z_c$. Here $z_l$ is the representation of language-switching behavior, which is generated conditioned on the context representation 
$z_c$, which captures syntactic and structural properties of a sentence. $\Wb$ is the observed sequence of words and $\Yb$ is the corresponding label sequence. 
Given our objective, a hierarchical VAE architecture is adapted for the basic formulation with suitable departures whenever required.
The next subsections will cover in details the inference model, generative model and prior.

\subsection{Encoder}
\label{sec:OurModel:Encoder}

Given observed labeled sentence $(\Wb, \Yb)$, our inference model $q_{\phi}$ defines a hierarchical probabilistic encoding $Z=(z_c, z_l)$ by first learning the content, structural embedding $z_c$ of the entire sentence. Using this embedding $z_c$ along with sequentially learned language label information the inference model $q_{\phi}$ encodes the latent switching pattern embedding $z_l$. Figure~\ref{fig:encoderdecoder}\,(a) illustrates the encoder.
We use two distinct RNN (LSTM) cells in the encoder,
$\rqc$ and $\rql$.
Their corresponding recurrent states are denoted
$\hqc$ and $\hql$.
Input token positions are indexed as $i = 0, 1, \ldots, I$.
The recurrence to estimate $z_c$ goes like this.
\begin{align}
 \text{We initialize} \quad \hqc_0 &= \vec0 \\
 \text{For $i = 1, \ldots, I$:} \quad
 \hqc_i &= \rqc(w_i, \hqc_{i-1}) \\
 \text{Finally,} \quad
 [\mu_{q,c}, \sigma_{q,c}] &= f_{q,c}(\hqc_I) \\
 \text{and then} \quad
 z_c \sim q_{\phi}(z_c|\Wb) &=
 \mathcal{N}(\mu_{q,c}, \diag(\sigma^2_{q,c}))
 \intertext{Next we estimate the encoding~$z_l$.}
 \text{We initialize} \quad \hql_0 &= z_c \\
 \text{For $i = 1, \ldots, I$:} \quad
 \hql_i &= \rql(y_i, \hql_{i-1}) \\
 \text{Finally,} \quad
 [\mu_{q,l}, \sigma_{q,l}] &= f_{q,l}(\hql_I) \\
 \text{and then} \quad
 z_l \sim q_\phi(z_l|z_c, \Yb) &=
 \mathcal{N}(\mu_{q,l}, \diag(\sigma^2_{q,l})). \\
 \text{Overall,} \quad
 q_{\phi}(Z | \Wb, \Yb ) &=
 q_{\phi}(z_{c} | \Wb) q_{\phi}(z_{l} | z_{c}, \Yb) \notag
\end{align}
Here, $\mu_{q,c}, \sigma_{q,c}$ are the mean and standard deviation for the context encoding and $\mu_{q, l}, \sigma_{q,l}$ are the mean and standard deviation for the switching behavior encoding distribution. $\Ncal$ denotes normal distribution. $\text{diag}(\cdot)$ represents a diagonal covariance matrix. $\fqc, \fql$ are modeled as feed forward stages, $\rqc, \rql$ are designed as recurrent units. We use the subscript $q$ (or $p$) to highlight if it belongs to encoder (or decoder).

Summarizing the distinction from traditional hierarchical VAE, \ourmodel's inference module accepts inputs in both encoding layers: word sequence at the ground layer and language label sequence at the upper layer. Learning a sequence model over language labels becomes difficult (even with hierarchical encoding) if we provide both the inputs only in the lowest level, possibly by concatenating suitable embeddings~\cite{winata2018code}.

\subsection{Decoder}
\label{sec:OurModel:Decoder}

Starting from $Z=(z_l, z_c)$, our probabilistic decoder generates synthetic code-switched text with per-token language ID labels, using a two-level hierarchy of latent encoding. However, unlike the conventional hierarchical variational decoder, \ourmodel{} decodes a switching pattern given $z_l$ at the upper level, then conditioned on $z_l$ and the decoded tag history it generates a content distribution $z_c$. Here we need to design a specific decoupling mechanism of $z_c$ from $z_l$, which is not just the reverse of encoding technique. As $z_l$ has the switching information as well as the context information, we use both $z_l$ and $h(y)$ which is the history of label decoding to decode the distribution of $z_c$. We design the loss in such a way that tries to minimize the difference between encoding and decoding distribution of $z_c$.

We use two distinct RNN (LSTM) cells in the decoder, $\rpl$ and
$\rpc$. Their corresponding recurrent states are denoted $\hpl$ and
$\hpc$. Output token positions are indexed $o=1,\ldots,O$. The
feedforward network to convert $\hpl_o$ to a multinomial distribution
over $y_o$ is called $\fpl$.
\begin{align}
 \text{We initialize} \quad
 \hpl_0 &= z_l \; \text{and} \; y_0 = \SOS \\
 \text{For $o = 1, \ldots, O$:} \quad
 \hpl_o &= \rpl(y_{o-1}, \hpl_{o-1}) \\
 y_o &\sim \text{Multi}(\fpl(\hpl_o))
\end{align}
Decoding continues until some $y_O = \EOS$ is sampled at some $O$, and
then stops. Effectively this amounts to sampling from
$p_\theta(\Yb|z_l)$. Once all labels $\Yb = y_1, \ldots, y_O$ are
generated, we decode $z_c$ and start generating words $w_1, \ldots, w_O$.
$\fpc$ denotes the feedforward network to 
decode $z_c$ as follows.
\begin{align}
 [\mu_{p,c}, \sigma_{p,c}] &= f_{p,c}(h^{p,l}_{O}, z_l)
 \\
 z_c \sim p_\theta(z_c|z_l,\hpl_O )&= \Ncal(\mu_{p,c},\text{diag}(\sigma_{p,c}^2))
\end{align}
The feedforward network $f_{p,w}$ converts
$\hpc_o$ to a
multinomial distribution over words from the languages indicated by
$y_1,\ldots,y_O$.
\begin{align}
 \text{We initialize} \quad
 \hpc_0 &= z_c \quad \text{and} \quad w_0 = \SOS \\
 \text{For $o=1,\ldots,O$:} \quad
 \hpc_o &= \rpc(w_{o-1}, \hpc_{o-1}) \\
 w_o &\sim \text{Multi}(f_{p,w}(\hpc_o, y_o))
\end{align}
If $y_o = s$, $\fpc$ returns a multinomial distribution over the
source vocabulary, and if $y_o=t$, $\fpc$ returns a multinomial
distribution over the target vocabulary. Effectively, we have sampled
$\Wb$ from the distribution $p_\theta(\Wb|\Yb, z_c)$. Overall,
decoding amounts to sampling from
$p_{\theta}(\Yb, \Wb | Z) = p_{\theta}(\Yb | z_{l}) \;
p_{\theta}(\Wb | \Yb,z_{c})$.

Figure~\ref{fig:encoderdecoder}~(b) illustrates the decoder architecture. \ourmodel{} departs from existing dual RNN architectures \cite{garg2018code} (two RNNs dedicated to $s$ and $t$) in the following two ways:
\begin{itemize}[leftmargin=*]
\item Instead of using a softmax output of two decoding RNNs corresponding to two language generators, \ourmodel{} learns to decode language labels explicitly from a latent space.
\item By using tightly-coupled decoding RNNs, parameter learning in \ourmodel{} becomes more effective.
\end{itemize}
This way the decoder can generate word sequence in a more controlled fashion. The recursive word decoding unit generates a word given the predicted label from the \textit{language ID} decoding layer.

\subsection{Prior}
The latent variable $z_l$ can be sampled from the standard normal distribution:
\begin{align}
p_\pi(z_{l}) &\sim \Ncal(\mathbf{0}, \mathbb{I})
\intertext{and then reuse $p_\theta(z_c|z_l)$ to define}
p_\pi(Z) &= p_\pi(z_{l})\, p_{\theta}(z_{c}|z_{l} )
\end{align}

\subsection{Training} 
Given a collection of $M$ code-switched text ${S^{(m)} =
(\Wb^{(m)}, \Yb^{(m)}): m=1,\ldots,M}$, we train our model by maximizing the evidence lower
bound (ELBO), as described in Section~\ref{sec:Background}. In our case, after taking into consideration the dependence between $z_c$ and $z_l$, the ELBO can be 
simplified to:
\begin{align}
\max_{\phi,\theta}
& \!\! 
\sum_{m\in[M]} \! \Bigl[\EE_{q_{\phi}(Z^{(m)} | \Wb^{(m)}, \Yb^{(m)}}
\log p_{\theta}(\Wb^{(m)}, \Yb^{(m)} | Z^{(m)}) \notag \\
& -\EE_{q_\phi(z_c| \Wb^{(m)})}\text{KL}(q_{\phi}(z_{l}| z_c, \Yb^{(m)}) ||
p_\pi(z_l)) \notag \\
&-\EE_{q_\phi(z_c| \Wb^{(m)}), p_\theta(z_c | z_l)}\text{KL}(q_{\phi}(z_c) || p_\theta(z_c| z_l)) \Bigr] \label{eq:obj-old}
\end{align}
Because human-labeled code-mixed text is scarce, we first train \ourmodel{} with the parallel corpora specified in Section~\ref{sec:traindata}, with aligned word embeddings. Then we further tune model parameters using real code-switched data, also specified in Section~\ref{sec:traindata}. We used Adam optimiser and KL cost annealing technique as described ~\cite{bowman2015generating} to train \ourmodel.

\section{Experimental setup}
\label{sec:ExptSetup}

Here, we describe the labeled data sets, baseline paradigms, and evaluation criteria, followed by the description of language models used to evaluate the utility of the synthesized text. 

\subsection{Data sets to train generative models}
\label{sec:traindata}

To train the generative models, we use a subset of the (real) Hindi-English tweets collected by \cite{patro2017all} and automatically language-tagged by \cite{rijhwani2017estimating} with reasonable accuracy. From this set we sample 6K tweets where code-switching is present, which we collect into folds \textbf{rCS-train}, \textbf{rCS-valid}.
5K tweets are found labeled with only one language. These monolingual instances are converted into parallel corpora by translating Hindi sentences to English and vise versa using Google Translation API\footnote{\protect\url{https://translation.googleapis.com}}, generating 10K instances. The word embeddings of the two languages are aligned.

\subsection{Baseline generative models}

\xhdr{Deep generative models}
To understand the difficulties of extending existing monolingual text generators to code-switched text, we design four baselines from two state-of-the-art generative models. \cite{bowman2015generating} showed impressive results at generating monolingual sentences from a continuous latent space. They extended RNNlms with a variational inference mechanism. However, their model does not allow inclusion of hand crafted features like language ID, POS tag etc. Meanwhile, \cite{zhang2017adversarial} proposed a GAN model to generate a diverse set of sentences. Based on these, our baseline approaches are:
\begin{description}[leftmargin=*]
\item[pVAE:] We train the network of \cite{bowman2015generating} with the parallel corpora. The probability of generating a word is designed as a softmax over the union vocabulary. As both of the corpora are mapped to the same latent space due to aligned embeddings, we expect the model to switch language whenever it finds a word from the other language more probable than a word in the same language as the current word.
\item[rVAE:] To further make the model learn specific switching behaviors we train the model with the real code mixed text along with the parallel data.
\item[pGAN:] Similar to pVAE, we train the network proposed by \cite{zhang2017adversarial} with the parallel text corpora. 
\item[rGAN:] GAN trained with the real code-switched data along with the parallel corpora.
\end{description}

\xhdr{RNNlm based generative models}
Though language models are built primarily to estimate the likelihood of a given sentence, they can also be used as a generative tool. Recently, RNN based language models have been used to generate code-switched text as well, giving significant perplexity reduction compared to generic language models that do not consider features specific to code-switching. We compare \ourmodel{} against the following code-switched LMs:
\begin{description}[leftmargin=*]
\item[{aLM}:] We train the system proposed by \cite{winata2018code} using the real code-switched text and then use a word decoder and language decoder to generate synthetic texts.
\item[{bLM}:] After training the system proposed by \cite{chandu2018language} with the real code-mixed text, we use their LSTM decoder to generate synthetic text.
\end{description}

\begin{figure}[th]
\centering
\includegraphics[trim={0 0 1mm 5mm}, clip, width=0.75\hsize]{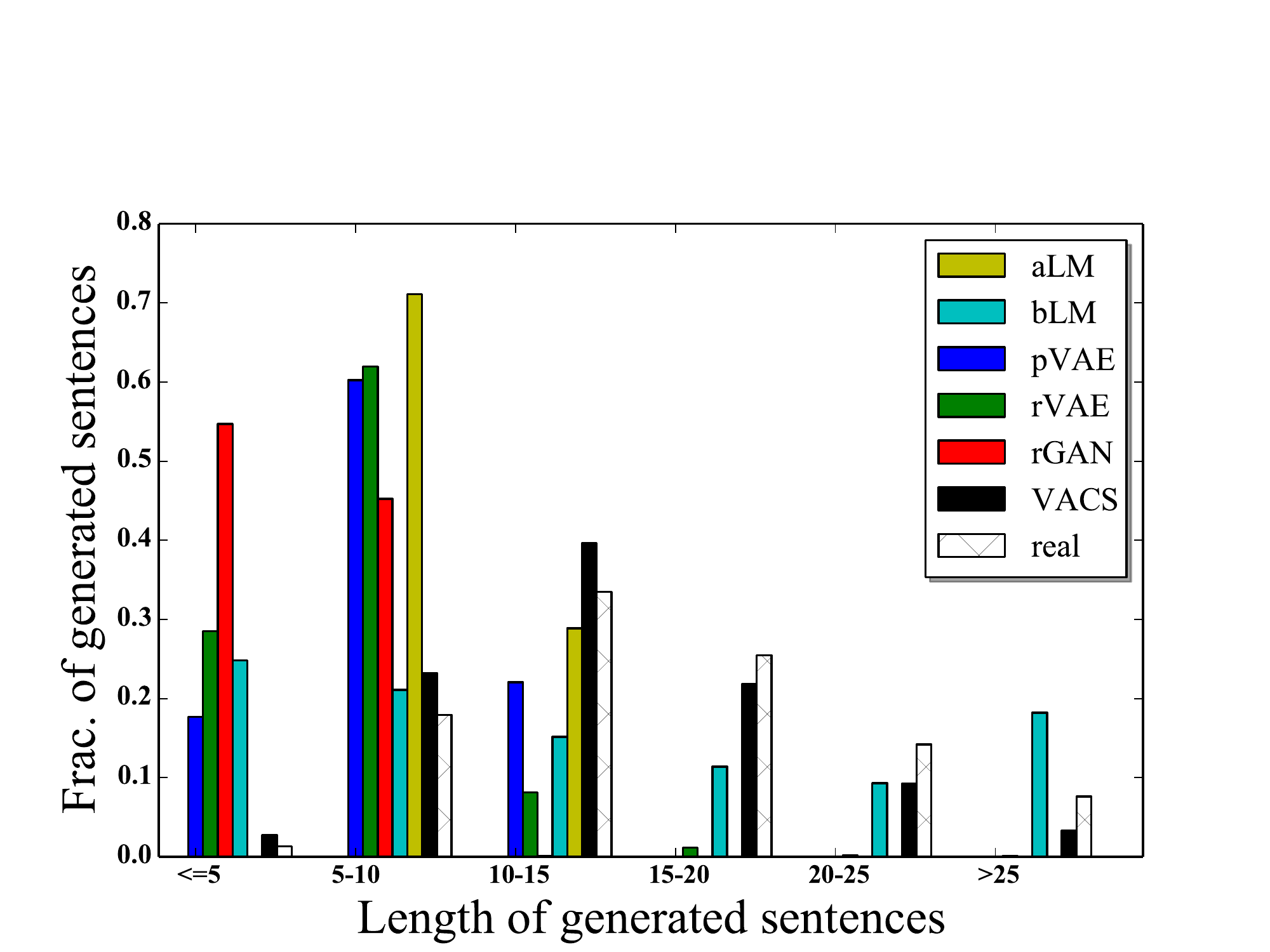}
\caption{Length distribution of the generated sentences from different methods. \ourmodel\ generates closest length distribution.
}
\label{fig:length}
\end{figure}

\subsection{Direct/intrinsic evaluation}

Here we analyse the features like length distribution and diversity of code-switching of generated synthetic texts. 
We also measured one sentence level metric \textbf{Code-Mixing Index (CMI)} coined by ~\cite{gamback2016comparing},
and three corpus level metrics
\textbf{Multilingual index (M-Index)}, \textbf{Burstiness} and \textbf{Span Entropy} that were introduced in ~\cite{guzman2017metrics} to demonstrate how different the generated texts are from the training corpus in terms of switching.

\subsection{Indirect/extrinsic evaluation}

We will use prior methods and \ourmodel{} to generate large volumes of code-switched text. These will be used to train a \emph{payload} language model (as distinct from the \emph{generative} model of \ourmodel{} and baselines) --- specifically, the character-level LSTM proposed by \cite{kim2016character}. Each training corpus will result in a trained payload model. The various payload models will then be used to calculate perplexity \cite{brown1992estimate} scores on a held-out natural code-switched corpus. The assumption is that the payload model with the smallest perplexity was trained by the `best' synthetic text.

\paragraph{Training curricula:}
\cite{baheti2017curriculum} show that language models perform better when trained with an interleaved \emph{curriculum} of monolingual text from both the participating languages, then ending with code-switched (CS) text, rather than randomly mixing them. We build the curriculum from the following corpora:
\begin{description}[leftmargin=*]
\item[{Mono}:] 2K monolingual Hindi and 2K monolingual English tweets were sampled from the dataset. We translated Hindi to English and vice versa and make a set of 8K tweets.
\item[{X-gCS}:] This is the generated synthetic data. We sampled 5K generated synthetic code-switched text from various generative models. Here \textbf{X} denotes the generative method, which is one of \{\textbf{pVAE, rVAE, pGAN, rGAN, aLM, bLM, \ourmodel}\}.
\end{description}
The specific curricula we use are:
\begin{description}[leftmargin=*]
\item[Mono,] which uses no synthetic data.
\item[gCS $|$ Mono,] first synthetic then parallel monolingual.
\item[Mono $|$ gCS,] first parallel monolingual, then synthetic.
\end{description}
Here $C_1|C_2$ denotes the sequence of corpora used to train the language model.
\begingroup 
Designing multi-task losses to guard against catastrophic forgetting is left for future work. 
\endgroup

\begin{table}[th]
\small
\centering
\scalebox{0.9}{
\begin{tabular}{|l|r|r| r| r|}
\hline
Dataset & Avg~CMI & M-index & Burstiness & Span Entropy \\ \hline
\hline
rCS-train &0.56 & 0.778& 0.232& 1.498\\ \hline
\hline
pVAE-gCS &-0.20&+0.108 & +0.081& -0.527 \\ \hline
rVAE-gCS &+0.01 & +0.216& +0.058& -0.375\\ \hline
\rowcolor{green!5}
\ourmodel-gCS &+0.08& +0.078 & +0.065& -0.192\\ \hline
aLM-gCS &+0.12 & +0.201& +0.036& -0.299 \\ \hline
bLM-gCS &+0.14& +0.155& +0.023& -0.287 \\ \hline
rGAN-gCS &+0.17 & +0.219& +0.081&-0.622 \\ \hline
\end{tabular}
}
\caption{Different code-switching metrics of real and generated code-switched text.}
\label{tab:diverse}
\end{table}

\paragraph{Validation and testing:}
We sample 7K instances from the original real code-switched pool for validation and 7K for testing.
These are considered as scarce evaluation resources and not used in payload training.


\section{Results and analysis}
\label{sec:Results}

In Section~\ref{sec:Res:Props} we compare intrinsic properties of synthetic texts generated from various models.
In Section~\ref{sec:Res:ExPerp} we compare perplexities of payload language models prepared from text synthesized by various generators. Finally, in Section~\ref{sec:Res:Samples}, we present anecdotes about generated text and its quality. pGAN fails to generate any appreciable rate of code-switching. Therefore, we refrain from considering pGAN any further.

\subsection{Intrinsic properties of synthesized text}
\label{sec:Res:Props}

Based on 5000 synthetic sentences sampled from different generative methods, we report the following properties.

\paragraph{Length:} We investigate the quality of generation methods in terms of variation in length and diversity. Figure~\ref{fig:length} depicts that \ourmodel{} can generate sentences of different sizes, unlike the other generative models which can only produce short texts. Other than bLM, all baselines tend to produce sentences shorter than 15 words. But Figure~\ref{fig:length} shows that real sentences have average length $\sim$16 and may be as long as 25 words. rGAN generates very short sentences, at most 5 words long, and pVAE and rVAE generate sentences with an average length of $\sim$10. For aLM and bLM average lengths are $\sim$10 and $\sim$12 respectively. \ourmodel{} has a mean of $\sim$17 and follows the distribution of real code-switched data most closely.

\paragraph{CMI, M-index, Burstiness, Span-Entropy:} We report the metric values of the generated corpus
and the real corpus in 
Table~\ref{tab:diverse}. 
\ourmodel{} is closest to the real corpus in terms of M-index and Span entropy, which indicates the ratio different language tokens in the generated sentences
and language span distribution is closer to the real data. Though \ourmodel\ produces a larger CMI and burstiness as it can produce sentences of different lengths and various switching patterns; its CMI is still smaller than GAN, aLM, and bLM and burstiness smaller than GAN and pVAE. 
GAN generates the highest CMI and burstiness indicating haphazard switching patterns.
On the other hand, pVAE produces lowest CMI and span entropy indicating that the generated sentences are ``almost monolingual'' or the language spans are equal in length. rVAE produces CMI very close to real and less diverse in terms of both switching and length distribution. 
Along with generating Burstiness with various switching patterns
GAN also has highest burstiness

\begin{table}[ht]
\begin{center}
\begin{small}
\begin{tabular}{|l|r|r|r|}
\hline
 & Training Curricula & Valid PPL & Test PPL \\ \hline
\rowcolor{LightGrey}
1 & Mono & 3034.251& 3123.827 \\ \hline
\hline
\rowcolor{red!3}
2a & aLM-gCS $\mid$ Mono & 3094.998 &  3179.039 \\ \hline
\rowcolor{red!5}
 & bLM-gCS $\mid$ Mono & 3051.042 & 3123.510 \\ \hline
 \rowcolor{red!9}
 & rGAN-gCS $\mid$ Mono & 3206.175 & 3298.085 \\ \hline
\rowcolor{green!4}
 & pVAE-gCS $\mid$ Mono &  2383.426 &  2337.617 \\ \hline
\rowcolor{green!8}
 & VACS-gCS $\mid$ Mono &2243.578 & 2296.533  \\
 \hline
 \hline
\rowcolor{red!5}
2b & Mono $\mid$ aLM-gCS & 3083.314 &  3189.905 \\ \hline
\rowcolor{green!4}
 & Mono $\mid$ bLM-gCS & 2829.149 & 2896.337 \\ \hline
\rowcolor{red!9}
 & Mono $\mid$ rGAN-gCS & 3015.820& 3069.263 \\ \hline
\rowcolor{green!4}
 & Mono $\mid$ pVAE-gCS & 2807.296&  2869.633 \\ \hline
\rowcolor{green!4}
 & Mono $\mid$ rVAE-gCS &  2418.342&  2493.023 \\ \hline

\rowcolor{LightGreen}
 & Mono $\mid$ VACS-gCS & 2081.774&  2090.781 \\ \hline
 
\end{tabular}
\end{small}
\end{center}
\caption{Perplexity of payload language model using different training curricula. \ourmodel\ achieves the lowest perplexity. Green: lower perplexity than Mono baseline; yellow and red: larger perplexity than Mono baseline (gray).}
\label{tab:main}
\end{table}

\subsection{Extrinsic perplexity}
\label{sec:Res:ExPerp}

Table~\ref{tab:main} provides a comparative study on the perplexity achieved on real validation and test CS text, after training a payload language model with different curricula spanning parallel monolingual (Mono) and synthetically generated CS (gCS) text.

Obviously, a payload language model that has seen only monolingual text when training will have large perplexity on held-out real CS text, which shows a diversity of switching behavior, in terms of both syntax structure near switches and the distribution of words used in switched segments. We expect that, in the absence of real CS text adequate to train the payload model, large volumes of synthetic text will help.

Surprisingly, this does not happen for aLM-gCS, bLM-gCS, and rGAN-gCS. Adding these texts to the monolingual baseline makes payload perplexity generally \emph{worse} and much worse in some cases, in particular, rGAN-gCS. VAE has better success. For the gCS$|$Mono curriculum, pVAE-gCS improves upon the baseline, but rVAE-gCS does \todo{conjecture why?} 
worse. On further investigation we found that, pVAE generates $\sim$80\% monolingual data, this contributes to the monolingual corpus which makes the training more coherent than mixing with code-switched data with low quality. 
For the Mono$|$gCS curriculum, both pVAE and rVAE perform worse than Mono.

In sharp contrast, \ourmodel-gCS achieves the best (smallest) perplexity in both curricula, and much smaller than the Mono baseline. This shows that synthetic text from \ourmodel{} can be used effectively to supplement parallel monolingual corpora. pVAE is the second best choice. 
\todo{even though rVAE can learn switching?}


GAN-based synthetic text performs poorly. 
pGAN fails to generate any plausible code-switched text as it does not get any real code-switch samples from the parallel corpora. rGAN performance is also worse than other generative models. Though \cite{zhang2017adversarial} avoided mode collapsing problems common in GANs, we observed that the problem prevailed for longer sentences ($>$10 words) when trying to train with small amounts of code-switched text. The problem persists because CS text is much more syntactically diverse than monolingual corpus, so training a GAN with a small number of real samples produces sub-optimal results.

The performance of aLM and bLM, while better than GANs, is far from \ourmodel. These LMs are designed explicitly for code-switched languages and require language-tagged data. Hence the generative power of such models strongly depends on the size and quality of tagged training data available. 

\begin{table*}
\centering
\begin{tabular}{ll}
\hline & Sentences \\ \hline
a & \begin{tabular}[c]{@{}l@{}}\textcolor{blue}{hara gaya} pakistan \textcolor{blue}{hamen logon ke} tweet \textcolor{blue}{rato karane}\\
\textcolor{green!60!black}{(Pakistan defeats us to stop people from tweeting)} \\
\textcolor{blue}{apane logon ko} batting \textcolor{blue}{upalabdh} series \textcolor{blue}{ke} \\
\textcolor{green!60!black}{(Batting series available to our people)} \\
cricket run \textcolor{blue}{banaake kiya} SA \textcolor{blue}{ke haar} \\
\textcolor{green!60!black}{(Defeated SA by making runs in cricket)}
\end{tabular} \\
\hline
b & \begin{tabular}[c]{@{}l@{}}\textcolor{blue}{ladakiyon} 20 assembly \textcolor{blue}{pratishat se}\\ 
\textcolor{green!60!black}{(Girls from 20 assembly percent)}\\
\textcolor{blue}{vichaar bhee bikree} that is 25 \textcolor{blue}{guna} assembly \textcolor{blue}{ka} \\
\textcolor{green!60!black}{(Justice is also sold, that is 25 times assembly)} \\ 
assembly against asia \textcolor{blue}{pradesh} irfaan \textcolor{blue}{teesree har} breaking\\
\textcolor{green!60!black}{(Assembly against asia irfaan's third deafeat was breaking)}
\end{tabular} \\
\hline
c & \begin{tabular}[c]{@{}l@{}}\textcolor{blue}{sarkar kee sthaapana} jawaharlal \textcolor{blue}{achchhee ki}\\
\textcolor{green!60!black}{(Government's establishment was done well by Jawaharlal)}
\\ normal \textcolor{blue}{bikree vaalee ek tha} smartphone\\ 
\textcolor{green!60!black}{(One smartphone was for normal sale)}\\
modi \textcolor{blue}{aye ke} break it in 54 wheels\\
\textcolor{green!60!black}{(Modi came and break it in 54 wheels)}
\end{tabular}\\
\hline
\end{tabular}
\caption{Sentences synthesized by \ourmodel{}. Each row corresponds to sentences sampled from a fixed context representation. The Blue segments are in Hindi. Green: English translation of CS text.
}
\label{tab:sentences}
\end{table*}

\subsection{Sample synthetic sentences}
\label{sec:Res:Samples}

Table~\ref{tab:sentences} shows sentences generated by \ourmodel{}. \todo{check} All sentences in a row block are sampled from the same context embedding $z_c$, and each row corresponds to a different~$z_l$. Note that the generated sentences seem to be able to produce a similar context. Like row (a) corresponds to cricket and (b) to assembly. 
It learns to produce meaningful phrases most of the cases which seem reasonable syntactically; however, semantics and pragmatics are not realistic, like in monolingual synthesis.

\section{Conclusion}
\label{sec:End}

We proposed \ourmodel, a novel variational autoencoder to synthesize unlimited volumes of language-tagged code-switched text starting with modest real code-switched and abundant monolingual text. We showed that \ourmodel{} generates text of various lengths and switching pattern. We also showed that synthetic code-switched text produced by \ourmodel{} can help train a language model that then has low perplexity on real code-switched text. We further demonstrated 
that we can generate reasonable syntactically valid sentences.
As \ourmodel{} can generate plausible language-tagged code-switched sentences, these can be used for various downstream applications like language labeling, POS tagging, NER etc. 
\section*{Acknowledgments}

B. Samanta is funded by Google India Ph.D. fellowship and the ``Learning Representations from Network Data" project sponsored by Intel. 
We would like to thank Dr. Dipanjan Das and Dr.  Dan Garrette of Google Research, New York for their valuable inputs.

\bibliographystyle{abbrv}
\bibliography{ijcai19,voila}

\begin{thebibliography}{10}

\bibitem{bahdanau2014neural}
D.~Bahdanau, K.~Cho, and Y.~Bengio.
\newblock Neural machine translation by jointly learning to align and
  translate.
\newblock {\em arXiv preprint arXiv:1409.0473}, 2014.

\bibitem{baheti2017curriculum}
A.~Baheti, S.~Sitaram, M.~Choudhury, and K.~Bali.
\newblock Curriculum design for code-switching: Experiments with language
  identification and language modeling with deep neural networks.
\newblock {\em Proceedings of ICON}, 2017.

\bibitem{bidishaACL}
Bidisha, N.~Samanta, S.~Ganguly, and Chakrabarti.
\newblock Improved sentiment detection via label transfer from monolingual to
  synthetic code-switched text.
\newblock 2019.

\bibitem{bowman2015large}
S.~R. Bowman, G.~Angeli, C.~Potts, and C.~D. Manning.
\newblock A large annotated corpus for learning natural language inference.
\newblock {\em arXiv preprint arXiv:1508.05326}, 2015.

\bibitem{bowman2015generating}
S.~R. Bowman, L.~Vilnis, O.~Vinyals, and Dai.
\newblock Generating sentences from a continuous space.
\newblock {\em arXiv preprint arXiv:1511.06349}, 2015.

\bibitem{brown1992estimate}
P.~F. Brown, V.~J.~D. Pietra, R.~L. Mercer, S.~A.~D. Pietra, and J.~C. Lai.
\newblock An estimate of an upper bound for the entropy of english.
\newblock {\em Computational Linguistics}, 18(1), 1992.

\bibitem{chandu2018language}
K.~Chandu, T.~Manzini, S.~Singh, and A.~W. Black.
\newblock Language informed modeling of code-switched text.
\newblock In {\em Proceedings of the Third Workshop on Computational Approaches
  to Linguistic Code-Switching}, 2018.

\bibitem{chorowski2015attention}
J.~K. Chorowski, D.~Bahdanau, D.~Serdyuk, K.~Cho, and Y.~Bengio.
\newblock Attention-based models for speech recognition.
\newblock In {\em NIPS}, 2015.

\bibitem{donahue2015long}
J.~Donahue, L.~Anne~Hendricks, S.~Guadarrama, M.~Rohrbach, S.~Venugopalan,
  K.~Saenko, and T.~Darrell.
\newblock Long-term recurrent convolutional networks for visual recognition and
  description.
\newblock In {\em IEEE CVPR}, 2015.

\bibitem{gamback2016comparing}
B.~Gamb{\"a}ck and A.~Das.
\newblock Comparing the level of code-switching in corpora.
\newblock In {\em LREC}, 2016.

\bibitem{garg2018code}
S.~Garg, T.~Parekh, and P.~Jyothi.
\newblock Code-switched language models using dual rnns and same-source
  pretraining.
\newblock {\em arXiv preprint arXiv:1809.01962}, 2018.

\bibitem{goodfellow2014generative}
I.~Goodfellow, J.~Pouget-Abadie, M.~Mirza, B.~Xu, D.~Warde-Farley, S.~Ozair,
  A.~Courville, and Y.~Bengio.
\newblock Generative adversarial nets.
\newblock In {\em NIPS}, 2014.

\bibitem{guzman2017metrics}
G.~A. Guzm{\'a}n, J.~Ricard, J.~Serigos, B.~E. Bullock, and A.~J. Toribio.
\newblock Metrics for modeling code-switching across corpora.
\newblock In {\em INTERSPEECH}, 2017.

\bibitem{kannan2017adversarial}
A.~Kannan and O.~Vinyals.
\newblock Adversarial evaluation of dialogue models.
\newblock {\em arXiv preprint arXiv:1701.08198}, 2017.

\bibitem{kim2016character}
Y.~Kim, Y.~Jernite, D.~Sontag, and A.~M. Rush.
\newblock Character-aware neural language models.
\newblock In {\em AAAI}, 2016.

\bibitem{kingma2013auto}
D.~P. Kingma and M.~Welling.
\newblock Auto-encoding variational bayes.
\newblock {\em arXiv preprint arXiv:1312.6114}, 2013.

\bibitem{muysken2000bilingual}
P.~Muysken, C.~P. D{\'\i}az, P.~C. Muysken, et~al.
\newblock {\em Bilingual speech: A typology of code-mixing}, volume~11.
\newblock Cambridge University Press, 2000.

\bibitem{myers1997duelling}
C.~Myers-Scotton.
\newblock {\em Duelling languages: Grammatical structure in codeswitching}.
\newblock Oxford University Press, 1997.

\bibitem{patro2017all}
J.~Patro, B.~Samanta, S.~Singh, A.~Basu, P.~Mukherjee, M.~Choudhury, and
  A.~Mukherjee.
\newblock All that is {E}nglish may be {H}indi: Enhancing language
  identification through automatic ranking of the likeliness of word borrowing
  in social media.
\newblock In {\em EMNLP Conference}, 2017.

\bibitem{Pratapa:2018:ACL}
A.~Pratapa, G.~Bhat, M.~Choudhury, S.~Sitaram, S.~Dandapat, and K.~Bali.
\newblock Language modeling for code-mixing: The role of linguistic theory
  based synthetic data.
\newblock In {\em ACL Conference}, 2018.

\bibitem{rezende2014stochastic}
D.~J. Rezende, S.~Mohamed, and D.~Wierstra.
\newblock Stochastic backpropagation and approximate inference in deep
  generative models.
\newblock {\em arXiv preprint arXiv:1401.4082}, 2014.

\bibitem{rijhwani2017estimating}
S.~Rijhwani, R.~Sequiera, M.~Choudhury, K.~Bali, and C.~S. Maddila.
\newblock Estimating code-switching on twitter with a novel generalized
  word-level language detection technique.
\newblock In {\em ACL Conference}, volume~1, 2017.

\bibitem{rudra2016understanding}
K.~Rudra, S.~Rijhwani, R.~Begum, K.~Bali, M.~Choudhury, and N.~Ganguly.
\newblock Understanding language preference for expression of opinion and
  sentiment: What do hindi-english speakers do on twitter?
\newblock In {\em EMNLP Conference}, 2016.

\bibitem{sonderby2016ladder}
C.~K. S{\o}nderby, T.~Raiko, L.~Maal{\o}e, S.~K. S{\o}nderby, and O.~Winther.
\newblock Ladder variational autoencoders.
\newblock In {\em NIPS}, 2016.

\bibitem{sutskever2014sequence}
I.~Sutskever, O.~Vinyals, and Q.~V. Le.
\newblock Sequence to sequence learning with neural networks.
\newblock In {\em NIPS}, 2014.

\bibitem{vinyals2015show}
O.~Vinyals, A.~Toshev, S.~Bengio, and D.~Erhan.
\newblock Show and tell: A neural image caption generator.
\newblock In {\em IEEE CVPR}, 2015.

\bibitem{winata2018code}
G.~I. Winata, A.~Madotto, C.-S. Wu, and P.~Fung.
\newblock Code-switching language modeling using syntax-aware multi-task
  learning.
\newblock {\em arXiv preprint arXiv:1805.12070}, 2018.

\bibitem{zhang2017adversarial}
Y.~Zhang, Z.~Gan, K.~Fan, Z.~Chen, R.~Henao, D.~Shen, and L.~Carin.
\newblock Adversarial feature matching for text generation.
\newblock {\em arXiv preprint arXiv:1706.03850}, 2017.

\end{thebibliography}

\end{document}